# Supervised Machine Learning Algorithm for Detecting Consistency between Reported Findings and the Conclusions of Mammography Reports

Alexander Berdichevsky[1], Mor Peleg[1], and Daniel L. Rubin[2]

[1]Department of Information systems, University of Haifa, Haifa, Israel, 3498838;

morpeleg@is.haifa.ac.il, sashberd@gmail.com

[2]Department of Radiology and Medicine (Biomedical Informatics Research), Stanford University

Medical School, Stanford, CA, 94305; rubin@med.stanford.edu

Corresponding authors:

Mor Peleg
Assoc. Prof. of Information Systems
University of Haifa, 3498838, Israel
Rabin Building, room 7047
Email: morpeleg@is.haifa.ac.il
URL: http://mis.hevra.haifa.ac.il/~morpeleg/
Phone: 972-4-824-9641
Fax: 972-4-828-8522

Daniel L. Rubin
Assistant Professor of Radiology and Medicine (Biomedical Informatics Research)
Stanford University
1201 Welch Road, P285
Stanford, CA 95405
Email: dlrubin@stanford.edu
URL: http://rubin.web.stanford.edu/
Phone: 650-723-9495
Fax: 650-723-5795

**Abstract**

**Objective.** Mammography reports document the diagnosis of patients' conditions. However, many reports contain non-standard terms (non-BI-RADS descriptors) and incomplete statements, which can lead to conclusions that are not well-supported by the reported findings. Our aim was to develop a tool to detect such discrepancies by comparing the reported conclusions to those that would be expected based on the reported radiology findings.

**Materials and Methods.** A deidentified data set from an academic hospital containing 258 mammography reports supplemented by 120 reports found on the web was used for training and evaluation. Spell checking and term normalization was used to unambiguously determine the reported BI-RADS descriptors. The resulting data were input into seven classifiers that classify mammography reports, based on their Findings sections, into seven BI-RADS final assessment categories. Finally, the semantic similarity score of a report to each BI-RADS category is reported.

**Results.** Our term normalization algorithm correctly identified 97% of the BI-RADS descriptors in mammography reports. Our system provided 76% precision and 83% recall in correctly classifying the reports according to BI-RADS final assessment category.

**Discussion.** The strength of our approach relies on providing high importance to BI-RADS terms in the summarization phase, on the semantic similarity that considers the complex data representation, and on the classification into all seven BI-RADs categories.

**Conclusion.** BI-RADS descriptors and expected final assessment categories could be automatically detected by our approach with fairly good accuracy, which could be used to make users aware that their reported findings do not match well with their conclusion.

# BACKGROUND AND SIGNIFICANCE

The purpose of an imaging report is to document a specialist interpretation of images and relate the findings to the patient's current clinical symptoms and signs in order to diagnose or contribute to the understanding of their medical condition [1]. The reports are often structured [2] into paragraphs describing the type of radiology done, clinical history, comparison (i.e., previous radiograms available for comparison), findings, and impression of the radiologist who composed the report, which provides a conclusion and advice on appropriate further investigation or management. Referring physicians are expecting that the radiology reports would be accurate, clear, complete, and timely [3-4]. However, many reports contain unclear statements and sometimes reports are incomplete. Lack of clarity is often the result of using non-standard terms. Incompleteness may result in a conclusion which is inconsistent with the reported findings, questioning the accuracy of the report. Non-clarity and incompleteness may lead to incorrect conclusion by other doctors, and even to the administration of inappropriate procedures to the patient [5-9].

One way to improve mammography reporting is to adopt standardized terminology for reporting these imaging studies. The American College of Radiology (ACR) developed a controlled terminology and coding system called Breast Imaging Reporting and Data System (BI-RADS) [10]. The BI-RADS lexicon specifies standardized terms ("BI-RADS descriptors") to describe the imaging features of abnormalities observed on mammograms. The BI-RADS lexicon also provides "BI-RADS final assessment categories" which define the degree of suspicion of malignancy based on the radiology findings. The BI-RADS descriptors, sanctioned by the ACR, are also known as *sanctioned terms*; all other terms, including synonyms, are referred to as *unsanctioned terms*. The conclusion of a radiology report includes a BI-RADS final assessment category (hereafter called "BI-RADS category"), which is a number between 0 and 6 that represents a categorized description of the likelihood of malignancy.

In order to facilitate entry of standard terms, such as BI-RADS descriptors, into electronic health records, structured reporting templates and macros [11-13] that offer users enumerated lists of standardized terms have been developed. However, such tools are not well-accepted by radiologists who feel burdened with long lists of standard terms on the one hand, and restricted in the ways in which they can express themselves, on the other hand. At present, despite the availability of BI-RADS, most radiology reporting systems do not enforce the use of BI-RADS sanctioned terms other than providing a means of specifying a BI-RADS category in the impression of reports.

## OBJECTIVE

We focus on the problem of identifying inconsistency between mammography reported findings and conclusions. We developed a supervised machine learning classifier that normalizes non-standard terms and classifies mammography reports according to their reported findings into BI-RADS categories. We call the application that we have developed MAmmography Smart System (MA.S.S). The classification problem we address is more complex that standard classification problems because BI-RADS categories represent a scale of probability ranges of suspicion. They are not sharp categories and even physicians have disagreements regarding how specific mammography reports should be classified. Hence our problem is a problem of fuzzy classification. Therefore we developed a novel approach for (a) inferring a generic/canonical representation of reports belonging to each BI-RADS category and then (b) measuring similarity of a new report to each of the BI-RADS category representation.

These machine-deduced grades could then be compared with the radiologist-assigned conclusion categories to detect inconsistency between reported findings and conclusion.

## MATERIALS AND METHODS

**Data set**

We obtained a deidentified data set from an academic hospital. Ethics committee approval was obtained for this data set, through the institution. The files contain 258 complete reports containing the findings, clinical history, and conclusions (impression with BI-RADS category). The reports, including their findings and conclusions were established by an expert mammographer and assumed to be correct. Table 1 provides the number of reports grouped by BI-RADS category. 75% of this data set was used for training the machine learning algorithms.

Our test set augmented the remaining 25% of the data set with 120 additional mammography reports that we found on the Web (see Appendix A for web sources) and from [2].

Table 1. Data set overview

| BI-RADS category given by radiologist | Total number of reports from the Stanford dataset [training, testing] | Number of reports from the web and [2] |
|---|---|---|
| 0 | 20 [16,4] | 11 |
| 1 | 21[17,4] | 15 |
| 2 | 21[17,4] | 14 |
| 3 | 112 [84,28] | 27 |
| 4 | 72 [54,18] | 17 |
| 5 | 4 [3,1] | 19 |
| 6 | 8 [6,2] | 17 |
| Total | 258 [197, 61] | 120 |

**Algorithms and data representation of the MA.S.S application**

Figure 1 provides an overview of the processing done by MA.S.S' algorithms. The application includes seven classifiers (one for each BI-RADS category) that were each built from the training set of reports belonging to the respective BI-RADS category. The process shown in Figure 1 was carried our separately to build each of these classifiers, and consists of the following steps. The first step is used for data pre-processing and steps 2-3 are used to build the classifier (data representation).

(1) Spelling and term normalization

(2) Text summarization

(3) Determination of common context (syntactic) structure for the reports belonging to a BI-RADS category

The reports belonging to the test set went through the same process of data pre-processing and representation. Thus each report was represented in the same data representation as that of the BI-RADS categories. Then, we performed the fourth step:

(4) Semantic similarity measurement to assess similarity between each report and those in each of the seven BI-RADS categories.

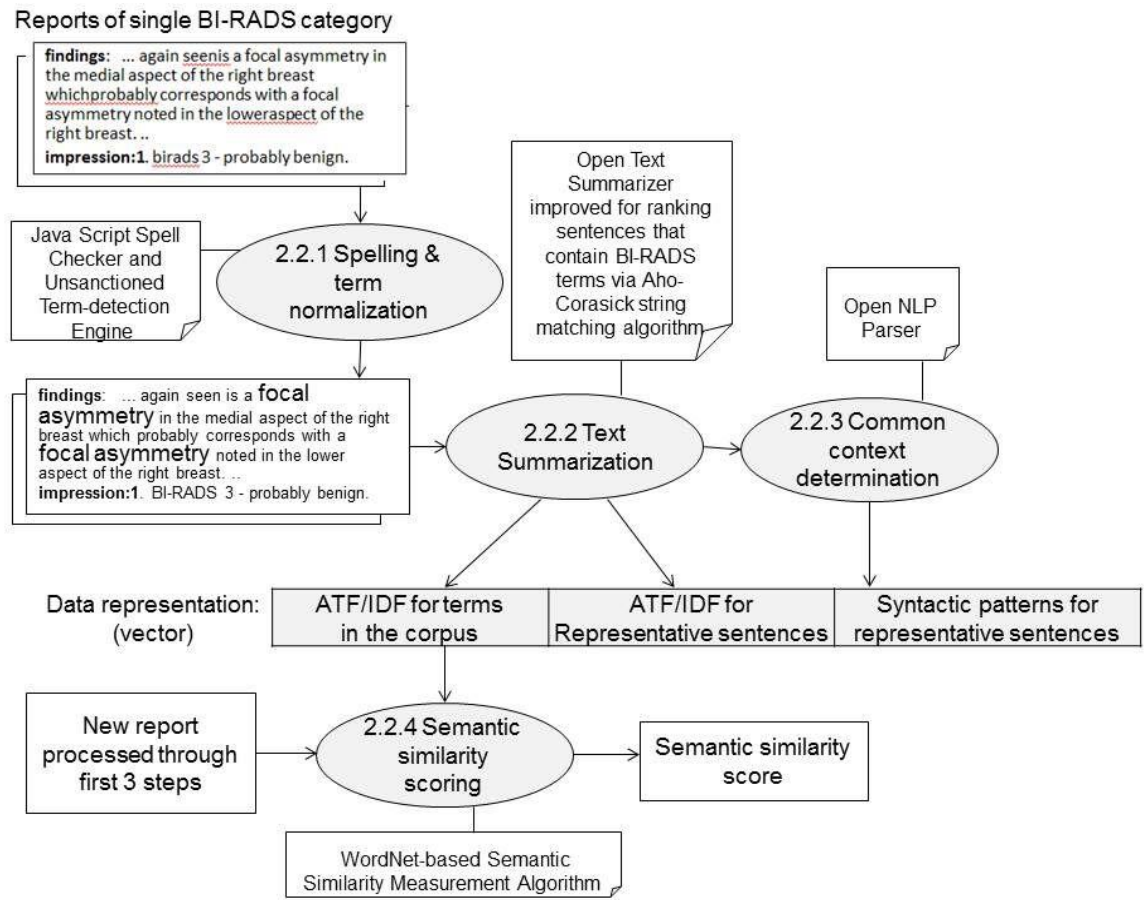

Figure 1. Preprocessing and classification methods of the MA.S.S application.

**Spelling and term normalization**

We implemented a Spell Checker and Unsanctioned Term-detection Engine to identify and propose corrections of spelling mistakes in the text. We used Java Script Spell Checker [14], which provides a simple and fast interface for identifying and fixing misspellings. Out tool has a user interface for interactive exploration: the spelling mistakes are shown and alternative words are suggested. Moreover, BI-RADS terms are recognized during typing (e.g., focal asymmetry, shown in Figure 1). Unsanctioned terms (e.g., "nodule") are recognized as specific misspelling by the interface, providing an option to users to exchange them with sanctioned terms (i.e., BI-RADS exact terms, e.g., "mass"). We envision that this tool would be used by radiologists as they are typing the report such that they would have a chance to normalize the terms. Therefore, we corrected the spelling mistakes and replaced non-sanctioned terms with BI-RADS exact terms before proceeding to the classification steps.

**Text Summarization**

The summarization step takes as input the findings section of a report belonging to a certain BI-RADS category. It computes the augmented term frequency (ATF) and inverse document frequency (IDF) for all input sentences and finds the set of the most important sentences for determining the BI-RADS category. We use the Augmented term frequency (i.e., raw term frequency divided by the maximum raw frequency of any term in the reports belonging to a certain BI-RADS category) to normalize according the size of the seven corpora (one per BI-RADS category), since in subsequent steps, the semantic similarity method consults the ATF and IDF scores of all BI-RADS categories.

The output of the summarization step is an XML structure that can schematically be represented as the multi-dimensional data structure shown in Figure 2. It records the ATF and IDF for all terms in the findings section of the report and stores the set of most representative sentences (see bottom of Figure 2), and for each one its ATF and AF.

During the training step, the output structure improves its accuracy of important sentences, which are selected from the growing corpus of reports belonging to a certain BI-RADS category.

Representing a selection of the most important sentences reduces the complexity of the algorithm, which in subsequent steps iterates through the detailed representation of the selected sentences. Even so, the representation vector for each classifier consists of over 300 nodes on average.

| | ATF/IDF for terms in the corpus | | | Representative sentences (centroid vector) | | | Syntactic patterns for representative sentences | | |
|---|---|---|---|---|---|---|---|---|---|
| | term | ATF | IDF | 1 2 3 4 5 6 7 8 9 10 11 12 | | | pattern/important word | #occurrences | locations |
| | seen | 0.001 | 1.2 | | | | [NP: DT JJ NN] | 22 | S1: 4, 6, 8, 13, 16, 18 ,… |
| | focal asymmetry | 0.02 | 1.27 | | | | [VP: VBZ JJ IN PRP$ JJ NN] | 4 | S2: 9, S3:8, … |
| | medial | 0.03 | 1.12 | | | | … | | |
| | aspect | 0.004 | 1.4 | | | | focal asymmetry | 30 | … |
| | breast | 0.01 | 1.8 | | | | | | |
| | … | | | | | | | | |

| sentence # | score | sentence text | ATF/IDF for the sentence | | | Syntactic patterns for the sentence | | |
|---|---|---|---|---|---|---|---|---|
| 1 | 20 | again seen is a focal asymmetry in the medial aspect of the right breast which probably corresponds with a focal asymmetry noted in the lower aspect of the right breast | term | ATF | IDF | pattern | #occurrences | locations |
| | | | seen | 0.124 | 1.222 | [NP: DT JJ NN] | 6 | 4 6 8,13,16,18 |
| | | | focal asymmetry | 0.011 | 1.647 | [ADVP: RB] | 1 | 1 |
| | | | medial | 0.052 | 1.291 | … | | |
| | | | aspect | 0.0227 | 1.758 | | | |
| | | | right | 0.19 | 1.194 | | | |
| | | | breast | 0.451 | 1.73 | | | |
| | | | … | | | | | |

Figure 2. Data representation for a BI-RADS category (the specific example is for BI-RADS category 3). The top part shows the representation for an entire BI-RADS category and the bottom for a representative sentence (out of the representative sentences of centroid vector).

To select the most important sentences for the BI-RADS category, the summarization algorithm determines for each sentence its heuristic score. The summarization algorithm that we have used is Open Text Summarizer Algorithm [15]. This algorithm is an improved free open-source version of Copernic Summarizer [16], which creates a concise summary of a document that highlights its most important concepts. The algorithm runs through given data, finds common citations, words, collocations etc. and then, using simple statistical tests, rates each data-sentence and presents its score signifying its degree of match to the BI-RADS category. However, the Open Text Summarizer algorithm did not pay any special attention to BI-RADS terms and its use did not produce a good summary of documents belonging to a single BI-RADS category. Therefore, we improved Open Text Summarizer with addition of Aho-Corasick pattern matching algorithm [17]. By integrating this algorithm into the Open Text Summarizer and configuring it to identify BI-RADS terms, we increased by a factor of two the score of sentences containing BI-RADS terms.

The number of representative sentences is configurable. We used the same number for all seven classifiers. We tested the performance of the MA.S.S system with 3 different numbers of representative sentences (5, 12, and 25) and used 12, which yielded the best results, which are reported in this paper.

**Determination of common context (syntactical) structure for a BI-RADS category**

One of the challenges in our research application was implementation of Natural Language Processing algorithms (NLP) in order to analyze the context (syntactic structure) of a given sentence and provide more exact classification of the input text. For this aim we used part of Open NLP Library [18] to create a Lexicography Tree for each sentence of the centroid vector produced by the text summarization step.

We stored two versions of the context structure for each sentence of the centroid vector: a string containing the sentence's words and their parts of speech, as shown in Figure 3 and a string containing only the parts of speech. For each representative sentence, the algorithm stores the number of occurrences of a given lexical pattern in the sentence and their locations. For example, the pattern [NP: DT JJ NN] appeared six times in locations 4, 6, 8, 13, 16, and 18 of the sentence shown in Figure 3. This pattern appears 22 times in representative sentences of the BI-RADS category=3, as shown in Figure 2). Moreover, we store the locations and count of for each of the important terms identified in the BI-RADS category (BI-RADS terms, and terms with high TF or with high IDF).

> "again seen is a focal asymmetry in the medial aspect of the right breast which probably corresponds with a focal asymmetry noted in the lower aspect of the right breast."
>
> "[ADVP again/RB ] [VP seen/VBN ] [NP is/VBZ ] [NP a/DT focal/JJ asymmetry/NN ] [PP in/IN ] [NP the/DT medial/JJ aspect/NN ] [PP of/IN ] [NP the/DT right/JJ breast/NN ] [NP which/WDT ] [ADVP probably/RB ] [VP corresponds/NNS ] [PP with/IN ] [NP a/DT focal/JJ asymmetry/NN ] [VP noted/VBN ] [PP in/IN ] [NP the/DT lower/JJ aspect/NN ] [PP of/IN ] [NP the/DT right/JJ breast/NN ] ."

Figure 3. Lexical structure (with words) for a sentence from a report of BI-RADS category of 3.

**Semantic similarity measurement**

Semantic similarity is assessed using the WordNet-based Semantic Similarity Measurement Algorithm [19]. This algorithm uses Word.Net lexical database which is available online and provides a large repository of English lexical items. The algorithm performs tokenization, word stemming, part of speech tagging, and word sense disambiguation. It then builds a semantic similarity relative matrix $R[m, n]$ of each pair of word senses (according to words in the Word.Net dictionary or to edit distance if no matching word is found), where $R[i, j]$ is the semantic similarity between the most appropriate sense of word at position i of X and the most appropriate sense of word at position j of Y. $R[i,j]$ is also the weight of the edge connecting from i to j. The problem of capturing semantic similarity between sentences is reformulated as the problem of computing a maximum total matching weight of a bipartite graph, where X and Y are two sets of disjoint nodes. The match results from are then combined (e.g., by averaging) into a single similarity value for two sentences.

Figure 4 shows the semantic similarity measure of a BI-RADS report to which the radiologist initially assigned BI-RADS category of 0. If the category matches the one that was inferred by the classifier then the user is notified and the report is added to the database. However, if the report matches other BI-RADS categories more closely, then, as shown on the bottom of the figure, the radiologist can see the similarity of the report to those in the various BI-RADS categories and he/she can then consider revising the BI-RADS category for this report by clicking on the "Submit" button below the highest-matching category (category 1 in the case of Figure 4).

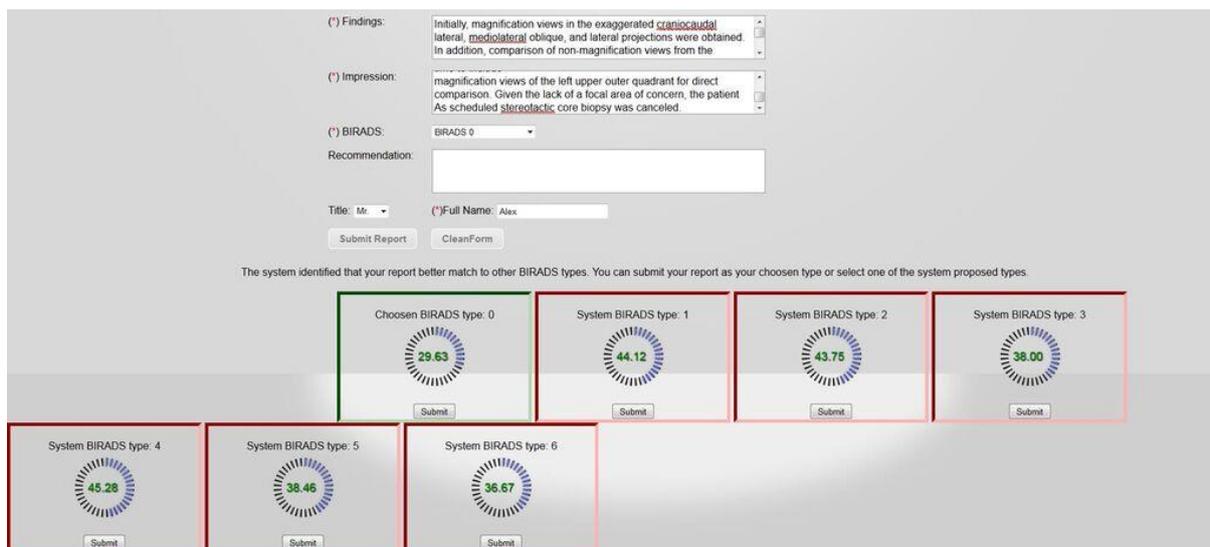

Figure 4. Output of the MA.S.S Application for a new report that is classified. The radiologist has given this report a BI-RADS category of 4 yet the classifier found this report to have a semantic similarity score of 43.4% with the BI-RADS 4, 50% parity to BI-RADS 1, and 45.83% to BI-RADS 2.

**Evaluation Methods**

Evaluation of the term normalization preprocessing step

Out of the reports contained in our dataset, 212 included occurrences of unsanctioned BI-RADS terms, which were determined by SQL queries complemented by manual inspection by the first author of this paper (see first column of Table 2). This was used as the gold standard for calculating precision and recall of BI-RADS terms that were identified by the normalization step.

**Classifier Evaluation**

As in the training set, each mammography report of our test set included a BI-RADS category that has been assigned by a radiologist. Each report from the test set went through summarization and context structure determining, creating a representative vector of the same structure as that of each BI-RADS classifier. Then the semantic similarity of each test report to each of the 7 BI-RADS category representations was measured. The BI-RADS category to which the similarity score was highest was selected as classifiers' output for the report. This was compared to the initial BI-RADS category that was assigned to the report by a radiologist (which was assumed to be correct as noted in the Data Set Section), and was taken as the gold standard.

To calculate precision and recall for each classifier we used the following definitions, where tp = true positives (i.e., report classified to BI-RADS category i when its original label was i); fp = false positives (i.e., report classified to BI-RADS category i when its original label was not i); fn = false negatives (report not classified to BI-RADS category i when its original label was i):

$$Precision\ (P) = \frac{tp}{(tp + fp)}$$

$$Recall\ (R) = \frac{tp}{(tp + fn)}$$

## RESULTS

**Results of evaluation of the term normalization preprocessing step**

Table 2 shows the results of evaluation of the term normalization algorithm. 206 (97%) of the unsanctioned BI-RADS terms were identified by the term normalization algorithm and the system correctly alerted the user in these cases that his/her report contained unsanctioned BI-RADS terms. The recall was 0.958 and precision was equal to 1.

Table 2. Evaluation results of the term normalization algorithm for unsanctioned terms

| Unsanctioned term | Number of occurrences in reports | Total identified (tp) | Total missed (fn) | Non-BI-RADS recognized as BI-RADS (fp) |
|---|---|---|---|---|
| density | 59 | 59 | 0 | 0 |
| vague density | 0 | 0 | 0 | 0 |
| nodule | 35 | 35 | 0 | 0 |
| ovoid | 13 | 13 | 0 | 0 |
| lobulated | 15 | 15 | 0 | 0 |
| poorly-defined | 0 | 0 | 0 | 0 |
| stellate | 2 | 2 | 0 | 0 |
| layering | 3 | 3 | 0 | 0 |
| teacup | 0 | 0 | 0 | 0 |
| tubular | 1 | 1 | 0 | 0 |
| tram-track | 0 | 0 | 0 | 0 |
| predominantly round | 4 | 2 | 2 | 0 |
| casting | 0 | 0 | 0 | 0 |
| indeterminate | 3 | 3 | 3 | 0 |
| heterogeneous | 70 | 69 | 1 | 0 |
| loosely grouped | 4 | 1 | 3 | 0 |
| ductal | 3 | 3 | 0 | 0 |
| Total: | 212 | 206(0.97) | 9(0.03) | 0 (0) |

**Results of evaluation of the classifier**

The results are reported in Table 3. On average, the system provides 76% precision and 83% recall in the classification task.

Table 3. Classifier precision and recall rates

| BI-RADS Category | Total Tested Reports | Total True Positives | Total False Positive | Total False Negative | BI-RADS category Precision | BI-RADS category Recall |
|---|---|---|---|---|---|---|
| 0 | 16 [5,11] | 7 | 5 | 4 | 0.58 | 0.63 |
| 1 | 20 [5,15] | 11 | 6 | 3 | 0.64 | 0.78 |
| 2 | 19 [5,14] | 4 | 9 | 6 | 0.30 | 0.4 |
| 3 | 58 [28,30] | 48 | 8 | 2 | 0.85 | 0.96 |
| 4 | 35 [18,17] | 25 | 5 | 5 | 0.83 | 0.83 |
| 5 | 20 [1,19] | 13 | 4 | 3 | 0.76 | 0.81 |
| 6 | 19 [2, 17] | 17 | 1 | 1 | 0.94 | 0.94 |
| Total: | 187 [64, 123] | 125 | 38 | 24 | 0.76 | 0.83 |

## DISCUSSION

Radiology reports convey the result of imaging study interpretations and are the primary communication vehicle between the radiologist and referring clinician. In mammography, the radiologists' assessment of likelihood of malignancy is conveyed by the BI-RADS category. Since the assignment of this category and the description of imaging features are independent events, it is possible that sometimes they are not concordant. The goal of our work was to detect such occurrences to enable better report quality, communication, and patient management.

We approached the problem as a term normalization and classification task. For the term normalization task, our methods achieved nearly perfect results with 97% of BI-RADS terms correctly identified. The precision and recall rates of the classifier were also good, despite the fact that the training corpus had only 193 reports (all from the institutional data set) and that the test set included in addition to the 65 reports from the institutional data set as well as reports from external sources. Training the system with a larger number of reports including reports from additional sources could result in better performance.

**Comparison with other medical report classifiers**

A number of other NLP systems have been developed for radiology, though only the first two described below developed classifiers specifically for the task of identifying the BI-RADS category of the report. Our work fills gaps present in the other classification approaches. The strength of our approach relies on the use of BI-RADS terms that receive high importance in the text summarization phase, similar to the utilization of BI-RADS by Nassif's classifier. Unlike related works, our approach includes a very detailed representation vector for each classifier; the Text Summarizer increases the score of sentences containing BI-RADS terms and the algorithm for determining Common Context Structure stores detailed information regarding BI-RADS terms (the locations and count of these terms) as part of the BI-RADS category's representative vector. The decision to provide higher importance and more detailed representation to BI-RADS terms has allowed us to form a good classifier. However, such model requires complex processing that were provided by the algorithms that we have integrated, which increases the complexity of the computations. In addition, our algorithm classifies the reports into a large number of categories (7). However, the precision and recall of our classifier (76% precision and 83% recall) is lower than that of the two other classsfiers.

The work that is most related to ours was done by Nassif et al. [20] who developed a general scheme for concept information retrieval from free text given a lexicon. They focused on improving feature-selection using a syntax analyzer, a BI-RADS concept finder, and a negation detector. Their algorithm achieved 97.7% precision and 95.5% recall. Their algorithm also allows the assessment of radiologist's labeling of mammography reports. By comparing the features extracted by the radiologist to the algorithm's output, they can detect repeatedly missed concepts and suggest areas for improvement. The difference between that work and ours is in the system's structure and in the approach to the users. Nassif's system is built of a set of rules and algorithms that extract important features for classifying the reports.

This is different than our more dynamic algorithms that remember each report belonging to the category and update the representation vectors to account for each added report, resulting in an explicit representation structure that can produce those sentences that are typical of a BI-RADS category, as done in our work. In addition, our system is client oriented; unlike Nassif's system, we do not force users to use standard terms when they create their reports. Instead we help users to be more exact in the description of findings by *suggesting* rather than forcing replacements for unsanctioned terms and pointing to missing descriptors of findings.

Percha et al. [21] also developed a classifier of mammography reports by breast tissue composition. Using a set of pattern-based rules, their algorithm assigns each report to one of four BI-RADS composition classes, while our classifier works for all seven BI-RADS categories. The method achieves >99% classification accuracy. To construct the rules they mapped all of the key terms and phrases from the full BI-RADS lexicon to specific breast composition classes. Working with experts they added additional terms (domain knowledge) that are characteristic of a breast composition class. They then mined the mammography report datasets for all words occurring in close proximity to the key terms and established rules for how far these words could reside before they ceased to be informative.

A simpler approach was used in [22]; commercially available software with embedded Boolean logic was used to train a classifier on radiology reports of ankle, spine and extremities, and categorize them into three categories: "fracture," "normal," and "neither", which may be simpler that classification into seven BI-RADS categories. The Boolean search terms were simple and related to the word "fracture", negation terms, and terms indicating generic abnormalities. The optimal proximity between words was determined empirically. Specificity for the three categories ranged from 91.6-98.1% and sensitivity ranged from 87.8-94.1%.

In line with the above works, and with our approach, which increases the importance of BI-RADS terms for report classification, Wilcox and Hripcsak [23] have shown that the role of domain knowledge is more important than the classification algorithm in improving classification results. They used the Medical Language Extraction and Encoding system (MedLEE) [24] to convert each report in their corpus of radiology reports and discharge summaries to a set of clinical observations. To prevent over-fitting, they limited attributes by their predictive value within the training set or by their medical relevance to the classification task as determined by domain knowledge. Specifically, they selected only relevant observations as determined from expert queries from the original studies evaluating MedLEE. Furthermore, they determined negations from the "certainty'' and ''status'' modifiers of terms extracted by MedLEE. In all of five classification algorithms that they have used, selecting features by relying on the medical relevance and negation yielded the best results.

Relying on additional medical knowledge such as patient demographic risk factors and radiologist-observed findings from consecutive clinical mammography examinations, allowed Burnside et al. [25] to develop a Bayesian Network classifier that can exceed radiologist performance in the classification of mammographic findings as benign or malignant. Similarly, Gupta et al. [26] used RadLex, a unified terminology for radiology, to enhance NLP-based mining for automated associative data mining of radiology text reports. Their algorithm mined the patient demographics with the findings in radiology text reports and was able to identify hemorrhage-related terms with specificity of 90.90% and sensitivity of 99.29%.

Chapman et al. [27] created a keyword classifier based of statistical properties of terms (similar to our text summarization phase) for automatic detection of chest radiograph reports containing findings consistent with inhalational anthrax. The classifier had specificity of 0.99 and sensitivity of 0.35.

Our work has several limitations. First, we do not know if better results could be achieved by other classification methods such as Artificial Neural Networks or Support Vector Machines. Our classifier was not developed using standard classification tools, but assembled a unique collection of algorithms. Our classifier is similar in a sense to K-Nearest Neighbor classification, because it makes explicit use of a user-defined similarity function (i.e., the user makes the final decision regarding which category to accept). In KNN, one is given a pool of many individual labeled examples; given a new example, its similarity to each of the known examples is computed, where the prominent label among the top most-similar K examples is assigned to the new example. Rather than using individual examples, we used a single processed representation per BI-RADS category, so only seven comparisons take place, and the scores of similarity to each category are presented to the user who decides which category to finally accept. The advantage of our approach over KNN is that we can handle the fuzzy classification problem that is inherent to BI-RADS categories, which represent probability ranges. In our approach, unlike KNN, we generate a feature vector of a canonical report belonging to a single BI-RADS score and we infer a similarity score between a new report and each representation of a BI-RADS category.

Another limitation of our work is that our classifier was trained on positive examples only. The precision of our classifiers could be potentially improved by training them also with negative examples consisting of all other available reports that have a different BI-RADS category. However, it remains an open question how to define the centroid vector for negative examples.

We could potentially improve the accuracy of the classifiers at the expense of complexity, by representing for each classifier all sentences in the corpus rather than representing just representative sentences.

Another potential way to improve classification is to improve the pre-processing algorithms. For example, our implementation did not include more complex algorithms (e.g., Yarowsky algorithm [28,29], collocation algorithm [30], Stanford compositional grammar parser [31]), although parts of those algorithms were implemented in our system. For example, our Summarizer algorithm identifiers common collocation, participle word sets, gerund etc., and performs statistical inference for them but on a very basic level. Our NLP engine also provides similar functionality to that of the Stanford Parser but with lower rates of accuracy, flexibility, and complexity. We are also interested in exploring the possibility of using NLP parsers that were especially developed to find longest noun phrases in biomedical domains and which work with medical vocabularies, such as Open-Biomedical Annotator (OBA) [32].

Our application could be used in other medical domains if trained with specific clinical context data. The general nature of this application is enhanced since the first three of its four algorithms are multi-lingual and could be used in other countries.

**CONCLUSION**

The MA.S.S application that we have developed explores new techniques for allowing radiologists to recognize unsanctioned terms and to be able to change them to standard BI-RADS terms, as well as to be aware of cases where there is a potential discrepancy between the reported findings and their reported BI-RADS category.

BI-RADS categories could be automatically deduced by our approach with fairly good accuracy, potentially enabling them to be compared with radiologist-reported categories to detect inconsistency between reported findings and their conclusion, as well as to permit assessing report completion.

**Acknowledgements.** We thank Dr. Einat Minkov for helpful discussion.

Appendix A – examples of web sources for mammography reports

1. ftp://medical.nema.org/medical/dicom/
2. http://medind.nic.in/ibn/t09/i4/ibnt09i4p266.htm
3. http://www.mtsamples.com/
4. http://training.seer.cancer.gov
5. http://www.supercoder.com/coding-newsletters/my-radiology-coding-alert/sample-report-challenge-test-your-mammogram-coding-skills-against-the-experts-article
6. http://www.dabsoft.ch/dicom/17/Q/
7. http://www.jabfm.org/content/19/2/161.full
8. http://www.bangkokmedjournal.com/sites/default/files/fullpapers/14-Nitida_AA.pdf